\pdfoutput=1
\documentclass[11pt,a4paper]{article}
\usepackage[hyperref]{eacl2021}
\usepackage{times}
\usepackage{latexsym}

\usepackage{url}
\usepackage{amsmath}
\usepackage{amssymb}
\usepackage{graphicx}
\usepackage[export]{adjustbox}
\usepackage{subcaption}
\usepackage{footnote}
\usepackage{multirow}
\makesavenoteenv{tabular}
\makesavenoteenv{table}
\usepackage{url}
\usepackage{tikz}
\usetikzlibrary{fit,positioning}
\usepackage{booktabs}

\usepackage{babel} % added by djamé

\DeclareMathOperator*{\argmax}{arg\,max}
\DeclareMathOperator*{\ELBo}{ELBo}
\DeclareMathOperator*{\IWAE}{IWAE}

\DeclareMathOperator*{\PIWO}{PIWO}
\DeclareMathOperator*{\iPIWO}{iPIWO}
\DeclareMathOperator*{\SSPIWO}{SSPIWO}
\DeclareMathOperator*{\SSiPIWO}{SSiPIWO}
\DeclareMathOperator*{\KL}{KL}
\usepackage{microtype}

\aclfinalcopy % Uncomment this line for the final submission
\usepackage{xcolor}

\expandafter\def\expandafter\UrlBreaks\expandafter{\UrlBreaks%  save the current one
  \do\a\do\b\do\c\do\d\do\e\do\f\do\g\do\h\do\i\do\j%
  \do\k\do\l\do\m\do\n\do\o\do\p\do\q\do\r\do\s\do\t%
  \do\u\do\v\do\w\do\x\do\y\do\z\do\A\do\B\do\C\do\D%
  \do\E\do\F\do\G\do\H\do\I\do\J\do\K\do\L\do\M\do\N%
  \do\O\do\P\do\Q\do\R\do\S\do\T\do\U\do\V\do\W\do\X%
  \do\Y\do\Z}

\usepackage{soulutf8}
\usepackage[normalem]{ulem}

\usepackage{enumitem}
\setlist{nolistsep}
\usepackage{microtype}

\iftrue % set to iftrue for camera ready version, iffalse to display draft comments
    \usepackage[final]{proofing}
  \renewcommand\hl[1]{{#1}}  %% to remove the highlith
   {\draftnote{\red{#2}}}
   \newcommand\redHL[1]{}
  \newcommand\todo[1]{}
  \newcommand{\Djame}[1]{}

\newcommand{\gfcmt}[1]{}
\newcommand{\gfcorr}[2]{}
\newcommand{\jlcmt}[1]{}
\newcommand{\jlcorr}[2]{}
\newcommand{\dscmt}[1]{}
\newcommand{\dscorr}[2]{}

\else  % the stuff below is for the draft mode (hl, comments and so)
\usepackage[draft]{proofing}

\newcommand{\gfcmt}[1]{\textcolor{orange}{#1}}
\newcommand{\gfcorr}[2]{\textcolor{orange}{#1 $\longrightarrow$ #2}}
\newcommand{\jlcmt}[1]{\textcolor{red}{#1}}
\newcommand{\jlcorr}[2]{\textcolor{red}{#1 $\longrightarrow$ #2}}
\newcommand{\dscmt}[1]{\textcolor{brown}{#1}}
\newcommand{\dscorr}[2]{\textcolor{brown}{#1 $\longrightarrow$ #2}}

\newcommand{\Djame}[1]{
\textbf{\textcolor{red}{\hl{Djame: #1}}}
}

\newcommand\red[1]{{\textbf{\textcolor{red}{#1}}}}

\let\oldred\red
\renewcommand\red[1]{{\bf \oldred{{#1}}}}

 \newcommand\redHL[1]{\red{\hl{#1}}}
\let\olddraftnote\draftnote
\renewcommand\draftnote[1]{\olddraftnote{\red{#1}}}

\fi

\title{Controlling the Interaction Between Generation and Inference in Semi-Supervised Variational Autoencoders Using Importance Weighting}

\author{Ghazi Felhi \\
  LIPN\\ Université Sorbonne Paris Nord\hspace{20px} \\ Villetaneuse, France \\
  \texttt{felhi@lipn.fr} \\\And
  Joseph Leroux \\
  LIPN\\ Université Sorbonne Paris Nord \\ Villetaneuse, France \\
  \texttt{leroux@lipn.fr} \\\And
  Djamé Seddah \\
  INRIA Paris \\ Paris, France \\
  \texttt{djame.seddah@inria.fr} \\}

\date{}

\begin{document}
\maketitle
\begin{abstract}
Even though Variational Autoencoders (VAEs) are widely used for semi-supervised learning, the reason why they work remains unclear. In fact, the addition of the unsupervised objective is most often vaguely described as a \emph{regularization}. The strength of this regularization is controlled by down-weighting the objective on the unlabeled part of the training set. Through an analysis of the objective of semi-supervised VAEs, we observe that they use the posterior of the learned generative model to guide the inference model in learning the partially observed latent variable. We show that given this observation, it is possible to gain finer control on the effect of the unsupervised objective on the training procedure. Using importance weighting, we derive two novel objectives that prioritize either one of the partially observed latent variable, or the unobserved latent variable. Experiments on the IMDB English sentiment analysis dataset and on the AG News topic classification dataset show the improvements brought by our prioritization mechanism and exhibit a behavior that is inline with our description of the inner working of Semi-Supervised VAEs.

%180 words
\end{abstract}

\section{Introduction}
Variational Autoencoders (VAEs) are a class of deep learning models that combine the efficiency of Amortized Variational Inference with the representational power of deep learning\cite{Zhang2019AdvancesInference}. They have been successfully applied to a range of Natural Language Processing (NLP) tasks such as text generation (\citealp{Bowman2016GeneratingSpace}; \citealp{Fang2019ImplicitGeneration}), multimodal generative modeling \cite{Shi2019VariationalModels}, disentangled representation learning (\citealp{John2019DisentangledTransfer}; \citealp{Chen2019ARepresentations}), and linguistically informed representation learning (\citealp{Wolf-Sonkin2018AInflection}; \citealp{Zhang2020Syntax-infusedGeneration}), \textit{inter alia}. Their successes can be attributed to their efficient learning process, their two-way mapping ability, and other properties such as the disentanglement they induce (\citealp{Rolinek2019VariationalAccident}; \citealp{Li2020ProgressiveRepresentations}; \citealp{Higgins2019-VAE:Framework}) and the resulting smoothness in their latent space \cite{Bowman2016GeneratingSpace}.\\
VAEs are also prevalently used for semi-supervised learning, as described in the pioneer work of \citet{Kingma2014a}. For that matter, the VAE's successes have been established through numerous works (\textit{e.g.} \citealp{Tomczak2018VAEVampprior}; \citealp{Chen2018VariationalLearning}; \citealp{Corro2019DifferentiableAutoencoder}), but to the best of our knowledge, their successes or failures are most often attributed to a \emph{regularization} (or \emph{over-regularization}) on the latent representations (\citealp{Chen2018VariationalLearning};  \citealp{Wolf-Sonkin2018AInflection}; \citealp{Yacoby2020FailureTasks})\\
In this work, we aim to deepen our understanding of the interaction between the generative model and the classification model in Semi-supervised VAEs. Our contributions can be summed up as follows: 
\begin{enumerate}
    \item After laying out the technical background (section \ref{BACKGROUND}), we clarify analytically that the improvements yielded by SSVAEs over sheer supervision comes from the generative model guiding the inference model with its posterior on the distribution of the partially observed latent varable (section \ref{SSEXPLAIN})
    \item We design a technique for focusing the learning process on either of the unobserved  or the partially-observed latent variable to improve the learning performance while still lower-bounding the log-likelihood (section \ref{DISSECTING})
    \item  We offer experimental evidence that  aligns with the our conception of the inner workings of SSVAEs, and that encourages the use of our prioritization mechanism in semi-supervised learning (section \ref{EXPERIMENTS}). 
\end{enumerate}
% Section \ref{BACKGROUND} lays out the technical background of our work, namely SSVAEs and the tighter log-likelihood lowerbound of Importance Weighted Autoencoders (IWAE) \cite{Burda2016ImportanceAutoencoders}. In section \ref{SSEXPLAIN}, we build an objective that guides the partially observed latent variable in SSVAEs with the posterior of a generative model, and we shed light on it's commonalities with the objective of a SSVAE. To study the usability of our understanding of SSVAEs, we build an estimator for the aforementioned objective, and other related quantities in section \ref{DISSECTING}, and formulate our claims about these quantities. Finally, in section \ref{EXPERIMENTS}, we support our claims through the analysis of experimental evidence.

\section{Related Works}
Semi-supervision with VAEs has received a lot of attention in the recent years. After the pioneer work of \citet{Kingma2014a} on image classification, the method was extended to more tasks such as morphological inflections \cite{Wolf-Sonkin2018AInflection}, controllable speech synthesis \cite{Habib2019Semi-SupervisedSynthesis}, Parsing \cite{Corro2019DifferentiableAutoencoder}, Sequential Labeling \cite{Chen2018VariationalLearning} and many more. Systems have also been tweaked in various manners to improve the learning performance. \citet{Tomczak2018VAEVampprior} uses a mixture of variational posteriors with pseudo inputs to prevent the unsupervised loss from causing over-regularization. As latent variables are stochastic, \citet{Zhang2019AdvancesInference} proposes using a deterministic ancestor of a single latent variable in a VAE to perform classification, and constrain this deterministic ancestor with an adversarial term to have it abide by the values of the random latent variable.  \citet{Gururangan2020VariationalClassification} introduces a low resource pretraining scheme using VAEs to transfer representational capabilities from in-domain unlabeled data, to downstream tasks using minimum computational power. In a broad sens, the current state of the art on the tasks we used for this study in semi-supervised learning \citet{Chen2020MixText:Classification} relies on a carefully crafted training scheme of representation mixing between labeled and unlabeled samples, and on contextual pretrained embeddings from \cite{Devlin2018}.\\
Our work is also in line with literature taking interest in building deep graphical models that relate observations to latent variables with known meanings. As such, \citet{Siddharth2017LearningModels} builds a general framework for training deep graphical models using VAEs, and provides techniques (also using importance weighting) to estimate the ELBo no matter the dependency structure of the graphical model.\\

\section{Background}
\label{BACKGROUND}
In the following, we will denote our observable data by $x$, our unknown latent variables by $z$, and the partially observed latent variable that we want to predict by $y$.
\subsection{Semi-Supervised Learning with VAEs}
\label{BACKGROUNDSSVAE}
We will take interest in the widely used form of SSVAEs, namely what is referred to as the M2 model from the work of \citet{Kingma2014a}. In the most general case\footnote{Contrary to the work of \citet{Kingma2014a}, we will not assume independence between $z$ and $y$ in our derivations}, our generative model is $p_\theta(x) = \int p_\theta(x|z,y)p_\theta(y, z) dy dz$. The loss function $\mathcal{J}^\alpha$ from this model is constructed as the summation between 3 losses\footnote{The original formulation mixed sums and expectations. We only use expectations here for uniformity, and because the 3 quantities are calculated as averages over batches in the implementation for them to be equally scaled.}: \\
\begin{multline}
    \mathcal{J}^\alpha = \mathbb{E}_{(x, y)\sim p^*(x, y)}\left[\mathcal{L}(x, y)\right]\\
    +\mathbb{E}_{x\sim p^*(x)}\left[\mathcal{U}(x)\right] \\ + \alpha .\mathbb{E}_{(x, y)\sim p^*(x, y)} \left[-\log q_\phi(y|x)\right] \label{Jeq}
\end{multline}

Where $p^*$ is the distribution of our true data, $q_\phi$ is the approximate posterior (encoder) on $y$ and $z$, and $p_\theta$ is our generative model (decoder) (with $p_\theta(z, y)$ being the prior on $z$ and $y$).
$\mathcal{L}$ and $\mathcal{U}$ are explicited as follows:\\
\begin{multline}
\log p_\theta(x, y) \geq \mathbb{E}_{z\sim q_\phi(z|x, y)}\Bigl[\log p_\theta(x|y, z) \\
+ \log p_\theta(y, z)- \log q_\phi(z|x, y)\Bigr]\\
     = - \mathcal{L}(x, y ) = \ELBo(x, y) \label{L1}
\end{multline}

\begin{multline}
\log p_\theta(x) \geq \mathbb{E}_{(y, z)\sim q_\phi(y, z|x)}\Bigl[\log p_\theta(x|y, z) \\
+ \log p_\theta(y, z)- \log q_\phi(y, z|x)\Bigr]\\
     = - \mathcal{U}(x) = \ELBo(x) \label{U1}
\end{multline}
The ELBo (Evidence Lower Bound) is a lower bound on the true log-likelihood \cite{Kingma2019AnAutoencoders}. In equation \ref{Jeq}, the addition of the last term (weighted by $\alpha$ ) may seem to be driven by the need for a classification loss, but it can actually be derived as part of the lower bound for $\log p_\theta(x, y)$ (together with $-\mathcal{U}$) if we consider a symmetric Dirichlet prior for the categorical distribution on $y$ with parameter $\alpha$ \cite{Kingma2014a}.  $-\mathcal{J}^\alpha$ is therefore a sound lower bound for $\log p_\theta(x)+\log p_\theta(x, y)$.\\
Notice here that $\mathcal{L}$ and $\mathcal{U}$ respect the following equalities (\textit{c.f.} Appendix \ref{ULDERIV}):
\begin{multline}
\mathcal{L}(x, y ) = \log p_\theta(x, y) \\
- \KL[q_\phi(z|x, y)||p_\theta(z|x, y)] \label{L2}
\end{multline}

\begin{multline}
\mathcal{U}(x) = \log p_\theta(x) \\
- \KL[q_\phi(y, z|x)||p_\theta(y, z|x)] \label{U2}
\end{multline}
 Where $\KL[.||.]$ is the Kullback-Leibler divergence. The forms above are useful in that they isolates the effect on the inference network to their second  term. In  fact  the  inference  network $q_\phi$ receives no learning signal from the the first terms. This will prove useful in identifying the effect of the unsupervised objectives on the inference network in section \ref{DISSECTING}. 
However, especially when working on NLP problems, one should be wary that the intersection between $\phi$ and $\theta$ may or may not be the empty set. For instance, a very popular and effective design choice for language models is the use of tied embeddings \cite{Press2017UsingModels}. By using the same embedding weights for the encoder and the decoder, language models gain some performance and become much more memory efficient.\\
Using equations \ref{U2} and \ref{L2}, $\mathcal{J}^\alpha$ can be written as follows:

\begin{multline}
    \mathcal{J}^\alpha = - \mathbb{E}_{(x, y)\sim p^*(x, y)}\Bigl[\alpha \log q_\phi(y|x) \\
     +\log p_\theta(x, y) - \KL[q_\phi(z|x, y)||p_\theta(z|x, y)]\Bigr]\\
    - \mathbb{E}_{x\sim p^*(x)}\Bigl[\log p_\theta(x)\\ - \KL[q_\phi(y, z|x)||p_\theta(y, z|x)]\Bigr] \\ \label{Jeq2}\end{multline}

\subsection{Importance weighted Autoencoders}
\label{IWAESEC}
Although VAEs are an ubiquitous generative modeling choice, they remain faulty in that they only lower bound the true marginal log-likelihood. In that sense, various subsequent works have brought tighter lower bounds to the true log-likelihood such as \cite{Masrani2019TheObjective}, \cite{Ding2019LearningSampling}, and \cite{Burda2016ImportanceAutoencoders}. For this study, we chose to make use of IWAEs \cite{Burda2016ImportanceAutoencoders} as it is a mature technique that found its way into more applicative works such as that of \citet{Shi2019VariationalModels}. For a use case involving two latent variables ($y$ and $z$), the IWAE lower bound can be written as follows: 
\begin{multline}
    \log p_\theta(x) \geq \mathcal{L}_k(x) = \\ 
    \mathbb{E}_{(y_1, z_1), ...,(y_k, z_k) \sim q_\phi(y, z|x)}\\
    \left[\log\frac{1}{k}
    \sum_{i=1}^k\frac{p_\theta(y_i, z_i, x)}{q_\phi(y_i, z_i|x)} \right] \label{IWAE}
\end{multline}
The particularity of this lower bound (when contrasted to that of a VAE), is that it is equal to ELBo when $k=1$, and reaches the exact log-likelihood as $k$ goes to infinity. In other words, it departs from equation \ref{L2} at $k=1$, and anneals the Kullback-Leibler divergence in it as $k$ goes to infinity. We will make use of the Semi-Supervised IWAE (SSIWAE) objective in the experimental section, which is simply obtained by replacing the ELBo in equations \ref{L1} and \ref{U1} by a tighter importance weighted lower bound. Implications will be discussed in section \ref{DISSECTING}.

\section{How Do Semi-Supervised VAEs Improve Upon Supervised Learning?} \label{SSEXPLAIN}
For semi-supervision, $z$ is often assumed to be independent from $y$ \cite{Corro2019DifferentiableAutoencoder}, but some works do build hierarchical inference processes between $y$ and $z$ \cite{Chen2018VariationalLearning}. To keep our work valid for both cases, we won't assume independence between $z$ and $y$ for this section.\\
In the process of jointly learning generation and supervised inference, we will first think about how the former can help the latter. Generation can be learned given only the samples $x$, while supervised inference needs both $x$ and $y$. 
It is useful here not to think about semi-supervision as two separate learning processes, but rather as a process where the supervised signal is composed of a classical sample-wise learning signal, as well as an overall coherence statistic that is being validated with regard to the generative model.
We could give rise to such statistics validation without biasing the classical supervised maximum likelihood learning objective in the following manner:
\begin{align}
     \phi &= \argmax_\phi \mathcal{L}(y|x;\phi) \\
     &= \argmax_\phi \mathbb{E}_{x, y \sim p^*(x, y)} \Bigl[\log q_\phi(y|x) \Bigr] \\
     &= \argmax_\phi \mathbb{E}_{x, y \sim p^*(x, y)} \Bigl[ \alpha \log q_\phi(y|x) \nonumber\\
     & - \KL[q_\phi(y|x)||p^*(y|x)]\Bigr] \label{eq1} 
\end{align}
\\
Where $\alpha$ is a positive weight. These equalities stand because they are equalities between the $\argmax$ of each quantity, and not the quantities themselves. In fact, for an expressive enough approximation family $q_\phi$, our approximation is maximized at $q_\phi=p^*$ nullifying the KL-divergence. As a result, for an expressive enough $q_\phi$ this addition will not bias our learning objective. Since the KL term samples $y$ from $q_\phi$ and not $p^*$, it can indeed be calculated as a statistic over only the samples of $x$ (an expectation over only $x$) as is required for the unsupervised part of semi-supervised learning objectives. Equation \ref{eq1} can therefore be written as follows:
\begin{multline}
     \argmax_\phi \mathbb{E}_{x, y \sim p^*(x, y)} \left[\log q_\phi(y|x) \right] -\\ \mathbb{E}_{x\sim p^*(x)} \left[\KL[q_\phi(y|x)||p^*(y|x)]\right] \label{eq2}
\end{multline}
Additionally, if the family of approximations $p_\theta$ also contains $p^*$, we can write equation \ref{eq2} as follows:
\begin{multline}
 \argmax_\phi \mathbb{E}_{x, y \sim p^*(x, y)} \Bigl[\alpha \log q_\phi(y|x) \Bigr] -\\ \mathbb{E}_{x\sim p^*(x)} \Bigl[\KL[q_\phi(y|x)||p_{\theta^*}(y|x)]\Bigr] \\
      s.t.\hspace{8 px}  \theta^* = \argmax_\theta \mathbb{E}_{x\sim p^*(x)} \Bigl[\log p_\theta(x)]\Bigr]\\
        = \argmax_\theta \mathbb{E}_{(x, y)\sim p^*(x, y)} \Bigl[\log p_\theta(x, y)]\Bigr]
\end{multline}
Consequently, given that we would like to learn a generative model to guide our supervised model, a joint estimator that takes the following form could be used :
\begin{multline}
 \argmax_\phi \argmax_\theta\\
 \mathbb{E}_{x, y \sim p^*(x, y)} \Bigl[\alpha \log q_\phi(y|x) + \log p_\theta(x, y)\Bigr] \\ +\mathbb{E}_{x\sim p^*(x)} \Bigl[  \log p_\theta(x)  
 - \KL[q_\phi(y|x)||p_{\theta}(y|x)]\Bigr] \label{SSPIWO}
\end{multline}

Noticeably, this formalism closely resembles that of $-\mathcal{J}^\alpha$ in equation \ref{Jeq2}. The difference is in the Kullback-Leibler terms. While the objective in equation \ref{SSPIWO} only minimizes Kullback-Leibler divergence between $q_\phi(y|x)$ and $p_{\theta}(y|x)$, equation \ref{Jeq2} includes a Kullback-Liebler between $q_\phi(y, z|x)$ and $p_{\theta}(y, z|x)$ and another between $q_\phi(z|x, y)$ and $p_{\theta}(z|x, y)$. As stated before, the form that $\mathcal{J}^\alpha$ takes in equation \ref{Jeq2} is interesting in that it isolates the effect on the encoder in its Kullback-Leibler terms. Accordingly, the effect of semi-supervision with VAEs on a supervised learning process stems from  these terms. The Kullback-Leibler terms, in both the objective we constructed in equation \ref{SSPIWO} and $\mathcal{J}^\alpha$ in equation \ref{Jeq2}, work towards  bringing the $y$ we sample from the approximate posterior closer to that of the true posterior, which is the guidance provided by the generative model. It is this guidance that we deem responsible for the improvement brought by semi-supervision over sheer supervision. However, the terms that influence the inference network in equation \ref{Jeq2}, when optimized, also bring together the $z$ samples from the approximate posterior and those from the true posterior. We find it important to study the difference between the effect of the objective from equation \ref{SSPIWO}, and that of $-\mathcal{J}^\alpha$. In that sense, we will present a tractable estimator for the objective in equation \ref{SSPIWO}.

\section{Rebuilding semi-supervised VAEs with importance weighting}
\label{DISSECTING}
The problematic terms in equation \ref{SSPIWO} are $\log p_\theta(x, y)$ and the second expectation. $\log p_\theta(x, y)$ can be lower-bounded by the IWAE objective when considering $(x, y)$ an observation and $z$ a latent variable:
\begin{multline}
    \log p_\theta(x, y) \geq \\
     \mathbb{E}_{z_1, ...,z_k \sim q_\phi(z|x, y)} \Bigl[\log\frac{1}{k}\sum_{i=1}^k\frac{p_\theta(x, y, z_i)}{q_\phi(z_i|x, y)}\Bigr]\\
     =\IWAE(x, y, k)\label{SSPIWODERIV1}
\end{multline}
The content of the second expectation in equation \ref{SSPIWO} can be lower-bounded by the following estimator that partially uses the IWAE formalism, making it the Partially Importance Weighted Objective (PIWO):\\
\begin{multline}
    \log p_\theta(x) - \KL[q_\phi(y|x)||p_\theta(y|x)] \\
     \geq \mathbb{E}_{y \sim q_\phi(y|x)}\Bigl[ \\
     \mathbb{E}_{z_1, ...,z_k \sim q_\phi(z|x, y)} \Bigl[\log\frac{1}{k}\sum_{i=1}^k\frac{p_\theta(x, y, z_i)}{q_\phi(y, z_i|x)}\Bigr]\Bigr]\\
     =\PIWO(x, k)\label{SSPIWODERIV2}
\end{multline}
Its derivation is explicited in Appendix \ref{SSPIWODERIVAPPENDIX}.
Therefore, This estimator lower-bounds the objective in equation \ref{SSPIWO}, and the true log-likelihoods as in $-\mathcal{J}^\alpha$:\\
\begin{multline}
\log p_\theta(x) + \log p_\theta(x, y) \geq \\
\mathbb{E}_{(x, y) \sim p^*(x, y)}\Bigl[\alpha\log q_\phi(y|x)\\
+ \IWAE(x, y, k)\Bigr]\\
+\mathbb{E}_{(x) \sim p^*(x)}\Bigl[\PIWO(x, k)\Bigr]  \\
= \SSPIWO(x, y, k) \label{SSPIWODERIV}
\end{multline}

This is equal to $-\mathcal{J}^\alpha$ when $k=1$ and reaches the objective in equation \ref{SSPIWO} asymptotically when $k$ goes to infinity. We will refer to the lower bound in equation \ref{SSPIWODERIV} as the Semi-Supervised PIWO (SSPIWO). The above formalism prioritizes having $y$ abide by the true posterior over having both $y$ and $z$ abide by it  like in Semi-supervised VAEs (equation \ref{Jeq2}). A question naturally arises here: Is it always better to prioritize guiding $y$ with the true posterior ?\\
In a setup where annotated data is abundant, one could be tempted to concentrate the learning effort on $y$. However, in a setup where labels are scarce, multiple scenarios are likely: \textit{i)} $y$ overfits the few examples at hand and needs guidance from the generative model \textit{ii)} Learning $y$ needs to be delayed until after a good generative model is learned by concentrating the learning effort on $z$
\textit{iii)} Due to the difference in size, the supervised data has far lower data complexity than the unsupervised data, which leads to the learning process being dominated by the generation part. \\
Case \textit{i)} needs an objective that can concentrate the learning effort on $y$ like SSPIWO. However, case \textit{ii)} needs to concentrate learning on $z$. In that sense, and using the analogue process, we derive a $z$ centered learning objective that is similarly equal to $-\mathcal{J}^\alpha$ at $k=1$ and reaches the following lower bound when $k$ goes to infinity (full derivation in Appendix\ref{SSiPIWODERIVAPPENDIX}):\\
\begin{multline}
\log p_\theta(x) + \log p_\theta(x, y) \geq \\
     \mathbb{E}_{(x, y)\sim p^*(x, y)}\Bigl[ \alpha \log q_\phi(y|x)+\\
     \log p_\theta(x, y) - \KL[q_\phi(z|x, y)||p_\theta(z|x, y)]\Bigr]\\
    + \mathbb{E}_{x\sim p^*(x)}\Bigl[\log p_\theta(x) - \KL[q_\phi( z|x)||p_\theta(z|x)]\Bigr]\\ \geq \SSiPIWO(x, y, k)  \label{SSiPIWO}
\end{multline}

We dub this objective the Semi-Supervised inverse Partially Importance Weighted Objective (SSiPIWO).\\
%As SSPIWO and SSiPIWO are prioritization mechanisms, we expect them to be especially effective when learning resources are limited (\textit{i.e.} when the network size is too small for both latent variables $y$ and $z$ to be properly learned). 
 Finally, case \textit{iii)} requires down-weighting the generative objective as a whole from the inference network's learning objective. Given that weighting down the whole generative objective (or raising $\alpha$) has proven to be an effective strategy in semi-supervision with VAEs, the importance weighting used in SSIWAE should also induce an improvement over vanilla SSVAEs. In fact, the SSIWAE objective, as stated in \ref{IWAESEC}, asymptotically anneals the Kullback-Leibler divergence between posterior approximations and the true posteriors, which we deem responsible for the generative modelling's influence on the supervised learning process.\\
We synthesize the claims we aim to study in the following hypotheses: \emph{H1} : "SSPIWO improves upon SSVAE and SSiPIWO in labeled data abundant regimes", %\emph{H2} : "The improvement brought by our partially importance weighted objectives is especially salient for insufficient network sizes",
\emph{H2} : "The effect of complete importance weighting (SSIWAE) is similar to that of down-weighting the generative objective's influence on the classifier using a higher $\alpha$", and \emph{H3} : "The prioritization importance weighting objectives SSPIWO brings a lever for improvement that is different than that of down-weighting the generative objective". \\
\section{Experiments}
\label{EXPERIMENTS}
\subsection{Experimental Setup}
\paragraph{Datasets}
Sequence classification tasks are a classical testbed for semi-supervised learning. Therefore, we use the following datasets:
\begin{itemize}
    \item The binary sentiment analysis dataset IMDB. This dataset consists in 25K supervised examples and 50K unsupervised examples. The task here is to try to classify the sentiment expressed in movie reviews into negative, or positive sentiments. \cite{Maas2011LearningAnalysis}
    \item the text classification dataset AG News \cite{Zhang2015Character-levelClassification}. This task is 4-way topic classification task. We use 32K labeled examples as was done in \cite{Xu2017VariationalClassification} and 64K unlabeled examples to obtain similar proportions to IMDB.
\end{itemize}
The supervised examples are divided into 5 splits. For all experiments, we report the average accuracy over 5 runs, swapping each time the split we use for validation (early stopping, and $\alpha$ selection). The unlabeled training set is always fixed.
\paragraph{Graphical Model}

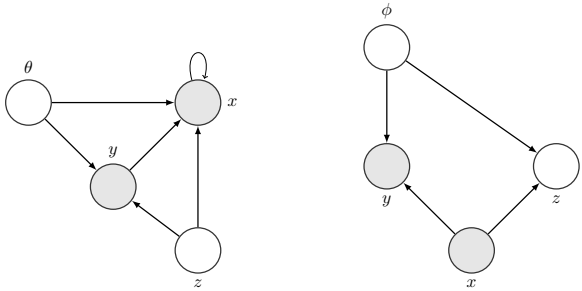
\begin{figure}[!h]
\centering
    \begin{minipage}[b]{0.5\textwidth}
    %\centering
    \begin{adjustbox}{minipage=\textwidth,scale=0.6}
        \begin{tikzpicture}
        \tikzstyle{main}=[circle, minimum size = 10mm, thick, draw =black!80, node distance = 16mm]
        \tikzstyle{connect}=[-latex, thick]
        \tikzstyle{box}=[rectangle, draw=black!100]
          \node[main, fill = black!10] (x) [label=below:$x$] { };
          %\node[main] (yemb) [above=of x,label=above:$y_{emb}$] { };
          \node[main] (z) [above right=of x,label=below:$z$] {};
          \node[main, fill = black!10] (y) [above left=of x,label=below:$y$] { };
          \node[main, fill = white!100] (phi) [above=of y, label=above:$\phi$] { };
          \path (phi) edge [connect] (z)
        		(phi) edge [connect] (y);
          \path (x) edge [connect] (y)
                (x) edge [connect] (z);
                
          \node[main] (z1) [,label=below:$z$, below left=of z, xshift=-60mm] {};
          \node[main, fill = black!10] (y1) [above left=of z1,label=above:$y$, yshift=-4.5mm%, xshift=-1.5mm
          ] { };
          \node[main, fill = black!10] (x1) [above right= of y1, label=right:$x$] { };
          \node[main, fill = white!100] (theta) [above left=of y1, label=above:$\theta$] { };
          \path (theta) edge [connect] (y1)
                (theta) edge [connect] (x1);
          \path (z1) edge [connect] (y1)
                (z1) edge [connect] (x1)
        		(y1) edge [connect] (x1)
        		(x1) edge [loop above] (x1);
          
        \end{tikzpicture}
    \end{adjustbox}
    \caption{Generative (left) and inference (right) Graphical Models}
    \label{fig:GRAPHMODS}
    \end{minipage}
\end{figure}

The graphical inference and generative models are depicted in figure \ref{fig:GRAPHMODS}. The white latent variables are Gaussian latent variables, while the grayed out ones are categorical. For the inference network, we model $y$ and $z$ as two conditionally independent latent variables (refer to Appendix \ref{STRUCTAPPENDIX} for a brief discussion of our objectives in the structured posterior case). In the generative model, we set the prior on $z$ to be a standard Gaussian, while $y$ is generated from a Categorical whose parameters are conditioned on $z$ through a learnable module. This design choice stems from the fact that we want the supervised $y$ in the inference model to be regularized by a distribution that depends on the example at hand. In fact, using independent priors would lead to regularizing $y$ with the marginal distribution of $y$ in the training set, while our modeling choice will provide $y$ with guidance that is specific to the current sample.\\
Finally, $x$ is generated auto-regressively using $y$, $z$, and its value at the previous time-step (the previous word).

\paragraph{Architectural choices and hyper-parameters}
In the inference (encoding) step, the Gaussian latent variables $z$ and $y$ are given by $q_\phi(z, y|x)=q_\phi(z|x)q_\phi(y|x)$. They are both obtained by passing $x$ through the same pretrained $300$-dimensional fastText\cite{Bojanowski2016} embedding layer, then taking the last state from a common 2-layered BiLSTM with $200$ hidden states. Their parameters are then obtained by passing this last state through 3 separate linear layers (1 for $\mu_z$, 1 for $\sigma_z$ and 1 for $logits_y$). $\sigma_z$ is obtained using an additional softplus activation gate. $z$ has size $100$ and the embeddings for $y$ have size $50$.\\
As for the decoding step, we model a structured prior $p_\theta(z, y)=p_\theta(y|z)p(z)$ where $p$ is a standard Gaussian and $p_\theta$ a linear layer yielding $logits_{y}$. The sampled $z$ and $y$ are then concatenated with the previous word at each generation step to obtain the next word using a 1-layered LSTM with size $200$.\\

% \begin{table}[t]
%     \centering
%     \resizebox{220px}{!}{
%     \begin{tabular}{|c||c|c|c|c|c|}
%     \hline
%     Model & $n_{emb}$ &  $n_{lstm1}$ &  $n_{z}$ & $n_{y}$ & $n_{lstm2}$   \\
%     \hline \hline
%     Experiment 1 & 300& 100& 200 &200 & 100 \\
%     Experiment 2 (n) & 20 & 20& 20& 10& 20 \\
%     \hline
%     \end{tabular}}
%     \caption{Network dimensions}
%     \label{NETSIZES}
% \end{table}

The experiments are conducted for the Semi-Supervised (SS) objectives "None" (pure supervision), "VAE", "PIWO", "iPIWO", and "IWAE". We try different supervision rates (percentages of supervised data with regard to total labeled training examples) ranging from 0.1\% to 100\%.\\
All experiments use a batch size of 32, a 0.5 dropout rate, and 4e-3 learning rate with the Adam optimizer \cite{Kingma2015}. For better efficiency, we reduce the vocabulary of the datasets to the 10000 most frequent words and we set the maximum sentence length to 256 tokens for IMDB and to 64 for AG News. No notable performance decay has been observed with these restrictions. \\
The stochasticity of the training procedure (random batches, latent variable samples, etc) brings variability in our experiments through phenomena that are out of the scope of this study. Nonetheless, we expect our hypotheses to be validated by the overall behavior of our results. The code for our experiments is publicly available\footnote{\href{https://github.com/ghazi-f/SSPIWO}{https://github.com/ghazi-f/SSPIWO}}.\\

\begin{table*}[t]
    \centering
    \resizebox{\textwidth}{!}{%
    \begin{tabular}{|c|c||c|c|c|c|c|c|c|}
    \hline
    dataset&SS & 0.1\% &  0.3\% &  1\% &  3\% & 10\% & 30\% & 100\% \\
    \hline \hline
    \multirow{5}{*}{IMDB} &None& 52.86\textcolor{gray}{(1.73)}& 51.73\textcolor{gray}{(1.71)}& 54.69\textcolor{gray}{(3.00)}& 56.19\textcolor{gray}{(4.14)}& 73.10\textcolor{gray}{(1.91)}& 82.48\textcolor{gray}{(0.55)}& 86.65\textcolor{gray}{(0.44)}\\ \cline{2-9}
    &VAE& 53.13\textcolor{gray}{(1.87)}& \textbf{53.99}\textcolor{gray}{(2.11)}& \textbf{57.58}\textcolor{gray}{(1.57)}& 61.47\textcolor{gray}{(1.38)}& 72.45\textcolor{gray}{(1.18)}& 82.44\textcolor{gray}{(0.85)}& 86.43\textcolor{gray}{(0.64)}\\ 
    &PIWO& 53.33\textcolor{gray}{(1.21)}& 53.91\textcolor{gray}{(2.57)}& 56.32\textcolor{gray}{(1.33)}& 60.81\textcolor{gray}{(1.79)}& \textbf{74.52}\textcolor{gray}{(1.30)}& \textbf{82.99}\textcolor{gray}{(0.54)}& \textbf{87.05}\textcolor{gray}{(0.50)}\\
    &iPIWO& 53.17\textcolor{gray}{(1.87)}& 52.70\textcolor{gray}{(1.69)}& 55.97\textcolor{gray}{(1.92)}& 59.71\textcolor{gray}{(2.84)}& 74.51\textcolor{gray}{(0.76)}& 82.56\textcolor{gray}{(1.11)}& 86.61\textcolor{gray}{(0.91)}\\ 
    &IWAE&\textbf{ 53.74}\textcolor{gray}{(1.54)}& 53.33\textcolor{gray}{(2.07)}& 56.28\textcolor{gray}{(2.42)}& \textbf{61.73}\textcolor{gray}{(1.36)}& 74.11\textcolor{gray}{(1.86)}& 82.32\textcolor{gray}{(0.74)}& 86.50\textcolor{gray}{(0.20)}\\ 
\hline
    \hline
    \multirow{5}{*}{AGNEWS}&None& 56.63\textcolor{gray}{(4.22)}& 68.49\textcolor{gray}{(5.69)}& 78.68\textcolor{gray}{(1.47)}& 82.82\textcolor{gray}{(2.24)}& 85.91\textcolor{gray}{(0.29)}& 88.38\textcolor{gray}{(0.48)}& 90.24\textcolor{gray}{(0.32)}\\  \cline{2-9} 
    &VAE& 60.23\textcolor{gray}{(2.54)}& 71.21\textcolor{gray}{(1.50)}& \textbf{78.73}\textcolor{gray}{(0.86)}& 82.14\textcolor{gray}{(0.96)}& 85.81\textcolor{gray}{(0.78)}& \textbf{88.26}\textcolor{gray}{(0.25)}& 90.13\textcolor{gray}{(0.29)}\\ 
    &PIWO& 61.83\textcolor{gray}{(4.41)}& 71.18\textcolor{gray}{(2.74)}& 78.55\textcolor{gray}{(1.12)}& 82.38\textcolor{gray}{(0.73)}& \textbf{86.10}\textcolor{gray}{(0.42)}&\textbf{ 88.26}\textcolor{gray}{(0.31)}& \textbf{90.24}\textcolor{gray}{(0.20)}\\ 
    &iPIWO& \textbf{63.01}\textcolor{gray}{(2.20)}& \textbf{72.21}\textcolor{gray}{(2.43)}& 77.24\textcolor{gray}{(1.68)}& 82.56\textcolor{gray}{(0.98)}& 85.79\textcolor{gray}{(0.69)}& 88.06\textcolor{gray}{(0.10)}& 90.03\textcolor{gray}{(0.35)}\\ 
    &IWAE& 62.40\textcolor{gray}{(3.22)}& 70.75\textcolor{gray}{(4.42)}& 77.82\textcolor{gray}{(2.17)}&\textbf{ 82.61}\textcolor{gray}{(0.93)}& 85.28\textcolor{gray}{(1.04)}& 87.99\textcolor{gray}{(0.44)}& 90.12\textcolor{gray}{(0.20)}\\ \hline

     \end{tabular}}
    \caption{Accuracies on IMDB and AGNEWS on their respective test sets.  The best \emph{semi-supervised} accuracy for each part of the table, and each supervision proportion is given in bold. The standard deviation for the 5 runs is between parentheses.}
    \label{tab:results1}
\end{table*}

\paragraph{Training Procedure}
\label{Training}
 To prevent posterior collapse, the kullback-leibler divergence is linearly annealed over the 3000 following steps as was done by \citet{Bowman2016GeneratingSpace}. For the multisample models (SSIWAE, SSPIWO, and SSiPIWO) we use 5 importance samples. We use the STL objective \cite{Roeder2017StickingInference} for ELBo, and the DReG objective \cite{Tucker2019DoublyObjectives} for importance weighted objectives, as these have been shown to lessen gradient variance without biasing the training procedure.\\
 If the accuracy doesn't improve for 4 epochs, the training is halted and the accuracy of the best epoch is reported.
 
\subsection{Results}

\begin{table}[t]
    \centering
    \resizebox{220px}{!}{
    \begin{tabular}{|c||c|c|c|c|}
    \hline
    dataset & SSVAE& SSPIWO &  SSiPIWO &  SSIWAE    \\
    \hline \hline
    IMDB & 1.86& 1.14& 1.86 & 1.29 \\
    AG News & 1.57 & 1.43& 2.0& 0.86 \\
    \hline
    \end{tabular}}
    \caption{Average $\log_{10}(\alpha)$ values across experiments}
    \label{tab:AlphaValue}
\end{table}
We compare the different objectives we are interested in, using different proportions of the labeled datasets. We picked the value for $\alpha$ that performed the best on the development set for each experiment in the set $\{10^{0}, 10^{1}, 10^{2}, 10^{3}\}$. The results are reported in table \ref{tab:results1}.\\
As stated in the previous section, the network sizes are fixed through all experiments. These sizes have been obtained by maximizing performance at 100\% of the data. Consequently, they are adequate for the High Supervision Rate (HSR) regime, and prone to overfitting for the Low Supervision Rate (LSR) regime.  For the upcoming analysis, we will refer to 0.1\% and 3\% as the LSR, and to 10\% to 100\% as the HSR.\\
\paragraph{HSR regime }In the HSR regime,  despite high variances, it can be seen that the the highest score is consistently obtained with SSPIWO. SSPIWO outperforming SSVAE and SSiPIWO constitutes evidence confirming hypothesis \emph{H1}. Additionally, SSPIWO improving upon SSVAE and SSIWAE in an experiment where we searched for $\alpha$ establishes that the benefit of its prioritization is complimentary to the benefit of varying $\alpha$ and thus distinct (as stated by hypothesis \emph{H3}). \\
\paragraph{LSR regime} As discussed in section \ref{DISSECTING}, there are different factors at play in the low supervision regime making it difficult for one objective to always outperform the others. This underlines the fact that even though SSPIWO is able to bring improvement, avoiding overfitting through careful architectural choices is still of great importance.\\
\paragraph{Studying the values of $\alpha$ }
A last observation we made in our experiments is reported in table \ref{tab:AlphaValue}, when inspecting the values of $\alpha$ that yielded the best performance for each objective. We can see that SSVAE selected values of $\alpha$ that were, in average, higher than those selected by SSIWAE by a large margin. This suggests that SSIWAE already down-weights the influence of the unsupervised objective to a certain extent, which is in line with Hypothesis \emph{H2}. Now looking at the prioritization objectives, we can also see  in both experiments that SSiPIWO has higher or equal $\alpha$ values than SSVAE, and that SSPIWO selects $\alpha$ values that are lower than those of SSVAE. Focusing on the partially observed latent variable rather than the unsupervised latent variable seems to allow for more learning signal from the unsupervised objective. This evidence further supports our claim about the core of the improvement brought by Semi-Supervised VAEs over purely supervised learning (section \ref{SSEXPLAIN}).
\paragraph{Relationship to multitask learning}
We consider it most interesting to remember that semi-supervision with generative models is a special case of multi-task learning where the performance on a main task is improved by jointly training for a proxy unsupervised task. Compatibility between multiple tasks in multi-task learning has been studied in numerous works. In the special case of NLP, the work of \citet{Binge2017IdentifyingNetworks} has shown that a major explanatory factor for the compatibility between main tasks and auxiliary tasks was whether gradients rose at similar moments during single task training.  Results from our work show that it is possible to alter the influence of the unsupervised proxy task in a sound manner so that a task that hindered learning the main task (\textit{c.f.} VAE vs None in table \ref{tab:results1} for 10\% to 100\% of IMDB ) becomes beneficial for it (\textit{c.f.} PIWO vs None results in the same setup). Therefore, importance weighting is a promising avenue of research to  find training schemes for multivariate VAE based graphical models that improve compatibility between main tasks and auxiliary tasks.

 \section{Discussion \& Conclusion}
Our work enabled better control on the learning process of semi-supervision with deep generative models. We conducted an analysis on the objective used in these models, which led to an alternative objective that could improve upon the vanilla SSVAE objective while still lower-bounding the true joint likelihood. The improvements brought by SSPIWO proved to be distinct from that of the weighting mechanism classically used in SSVAEs. We stated 3 Hypotheses in the theoretical section, which were in line with the the results provided by our experimental section. We now know that SSVAEs guide a partially observed variable by having it abide by their true posterior. We also know of a way to prioritize this guidance over learning other variability factors for the generative model, or conversely, how to prioritize factorizing the generative model into unknown factors while putting a minimum effort into aligning these factors with information from supervised variables. SSPIWO enables better injection of information about a known variability factor into a latent variable in regimes where labeled data is highly available. VAEs possessing a latent variable with a specific meaning are useful for semi-supervision, but also yield a model where the inference network can better explain the observations, and where the generative model can perform generation conditioned on specific values of the known factor. For suitably sized networks, SSPIWO could therefore improve such systems.\\
Although our study has brought to light new information about SSVAEs, it is limited with regard to two aspects. The first is that our new objectives use importance weighting, which linearly increases the memory cost of training. This disables the use of higher batch sizes for hardware with limited memory, or requires resorting to techniques such as gradient accumulation to lighten the memory constraints at the cost of a longer run time. The second limit is that this study focuses only on improving the classification score, while disregarding the generation score. We find it important for the whole joint likelihood lower-bounded by our objectives to be well estimated, leading to a model that can properly estimate labels from observation and conditionally generate observations on given labels. As such, we strongly advocate for the investigation of partially importance weighted objectives for conditional generation in future works. 
% \section*{Acknowledgments}

% The acknowledgments should go immediately before the references. Do not number the acknowledgments section.
% Do not include this section when submitting your paper for review.

\bibliography{references}
\bibliographystyle{acl_natbib}

\appendix

\section{Derivations for equations \ref{U2} and \ref{L2}}
\label{ULDERIV}
The following equality is known to be valid for an ELBo featuring $x$ as an observation and $z$ as a latent variable \cite{Kim2019VariationalAbstraction}:
\begin{multline}
\mathbb{E}_{z\sim q_\phi(z|x)}\Bigl[\log p_\theta(x|z) \\
+ \log p_\theta(z)- \log q_\phi(z|x)\Bigr] = \\
\log p_\theta(x) - \KL[q_\phi(z|x)||p_\theta(z|x)] \label{ELBoEQUALITY}
\end{multline}
Equation \ref{U2} (resp. equation \ref{L2}) is a direct application for this equality in a case where the observation is $x$ (resp. $(x, y)$) and the latent variable is $(y, z)$ (resp. $z$).

\section{Partially Importance Weighted Objective Derivation}

\label{SSPIWODERIVAPPENDIX}
We depart from the ELBo expressed considering only $y$ as a latent variable. The property exhibited by ELBo in equation \ref{ELBoEQUALITY} becomes as follows:
\begin{multline}
    \log p_\theta(x) - \KL[q_\phi(y|x)||p_\theta(y|x)] \\
    = \mathbb{E}_{y \sim q_\phi(y|x)}[\log p_\theta(x|y)] 
     - \KL[q_\phi(y|x)||p_\theta(y)]\\
    \label{SSPIWODERIV1}
\end{multline}
The left hand side is the content of the second expectation in equation \ref{SSPIWO} that we would like to estimate. The second term of the right hand size is tractable, but the first term is intractable. In fact, it would require marginalizing $p_\theta(x|y, z)p(z)$ over all possible values of $z$, and sampling from $p(z)$ is too unefficient for this to be done. Here we simply used the IWAE objective to lower-bound the first term (log-likelihood of $x$ conditioned on $y$) :

\begin{multline}
   \mathbb{E}_{y \sim q_\phi(y|x)}[\log p_\theta(x|y)]
     \geq\\ \mathbb{E}_{y \sim q_\phi(y|x)}\Bigl[ \\
     \mathbb{E}_{z_1, ...,z_k \sim q_\phi(z|x, y)} \Bigl[\log\frac{1}{k}\sum_{i=1}^k\frac{p_\theta(x, z_i|y)}{q_\phi(z_i|x, y)}\Bigr] \Bigr] \label{SSPIWODERIV2}
\end{multline}
Inequality \ref{SSPIWODERIV2} turns to an equality when $k$ goes to infinity. Replacing the first term in the right hand side of equation \ref{SSPIWODERIV1} with its lower bound from equation \ref{SSPIWODERIV2} leads to the PIWO lower bound :

\begin{multline}
    \log p_\theta(x) - \KL[q_\phi(y|x)||p_\theta(y|x)] \\
     \geq \mathbb{E}_{y \sim q_\phi(y|x)}\Bigl[ \\
     \mathbb{E}_{z_1, ...,z_k \sim q_\phi(z|x, y)} \Bigl[\log\frac{1}{k}\sum_{i=1}^k\frac{p_\theta(x, z_i|y)}{q_\phi(z_i|x, y)}\Bigr] \Bigr]\\ 
     - \KL[q_\phi(y|x)||p_\theta(y)] =\\
     \mathbb{E}_{y \sim q_\phi(y|x)}\Bigl[ \\
     \mathbb{E}_{z_1, ...,z_k \sim q_\phi(z|x, y)} \Bigl[\log\frac{1}{k}\sum_{i=1}^k\frac{p_\theta(x, y, z_i)}{q_\phi(y, z_i|x)}\Bigr]\Bigr]\label{SSPIWODERIV3}
\end{multline}
The last equality just integrates the decomposed Kullback-Leibler term into the expectation to make it more compact.

\section{Semi-Supervised Inverse Partially Importance Weighted Objective Derivation}
\label{SSiPIWODERIVAPPENDIX}
The quantity we aim to estimate is the following:
\begin{multline}
\SSiPIWO(x, y, k) \leq \\
     \mathbb{E}_{(x, y)\sim p^*(x, y)}\Bigl[ \alpha \log q_\phi(y|x) +\\
    \log p_\theta(x, y) - \KL[q_\phi(z|x, y)||p_\theta(z|x, y)]\Bigr]\\
    + \mathbb{E}_{x\sim p^*(x)}\Bigl[\log p_\theta(x)\\ - \KL[q_\phi( z|x)||p_\theta(z|x)]\Bigr]\\   \label{SSiPIWOAPPEN}
\end{multline}
The first expectation is clearly the same as in $\mathcal{J}^\alpha$ and will be trivially estimated using $\mathcal{L}(x, y)$ and $q_\phi(y|x)$.\\
For the term in the second expectation, we can use the exact derivation we used for PIWO while swapping the role of $y$ and $z$. In fact, this quantity is subject to the equality:

\begin{multline}
    \log p_\theta(x) - \KL[q_\phi(z|x)||p_\theta(z|x)] \\
    = \mathbb{E}_{z \sim q_\phi(z|x)}[\log p_\theta(x|z)] 
     - \KL[q_\phi(z|x)||p_\theta(z)]\\
    \label{SSiPIWODERIV1}
\end{multline}
It can thus be analogously lower-bounded by:
\begin{multline}
    \log p_\theta(x) - \KL[q_\phi(z|x)||p_\theta(z|x)] \geq \\
    \mathbb{E}_{z \sim q_\phi(z|x)}\Bigl[ \\
     \mathbb{E}_{y_1, ...,y_k \sim q_\phi(y|x, z)} \Bigl[\log\frac{1}{k}\sum_{i=1}^k\frac{p_\theta(x, y_i, z)}{q_\phi(y_i, z|x)}\Bigr]\Bigr] \\
     = \iPIWO(x, k) \label{SSiPIWODERIV2}
\end{multline}
The full estimator for SSiPIWO is therefore
\begin{multline}
\SSiPIWO(x, y, k) = \\
     \mathbb{E}_{(x, y)\sim p^*(x, y)}\Bigl[\alpha \log q_\phi(y|x) \\
    - \mathcal{L}(x, y)\Bigr]\\
    + \mathbb{E}_{x\sim p^*(x)}\Bigl[\iPIWO(x, k)\Bigr]\\   \label{SSiPIWOFINAL}
\end{multline}

\section{About Structured Approximate Posteriors}
\label{STRUCTAPPENDIX}
Modeling latent variables in the approximate posterior as independent latent variables is a choice that is most often sufficient. This ensues from the fact that, in most cases, observations are completely informative on the latent variables. Nevertheless, PIWO as can be seen in equation \ref{SSPIWODERIV3} requires sampling from $q_\phi(z|x,y)$ and, as a consequence, can only be computed exactly in cases where our variational posterior factors as $q_\phi(y, z|x)=q_\phi(z|x, y)q_\phi(y|x)$ (which includes the independence case). Otherwise, a lower bound to PIWO may be derived using a procedure similar to that used in \cite{Siddharth2017LearningModels} in section 2.1.\\
Of course, the analogous argument stands for iPIWO when swapping $z$ and $y$.
.

\end{document}